\documentclass[runningheads,hidelinks]{llncs}
\pdfoutput=1
\usepackage{booktabs} 
\usepackage{caption} 
\usepackage{subcaption} 
\usepackage{amsfonts} 
\usepackage{graphicx}
\usepackage{pgfplots}
\usepackage[all]{nowidow}
\usepackage[utf8]{inputenc}
\usepackage[english]{babel} 
\usepackage{tikz}
\usetikzlibrary{er,positioning,bayesnet}
\usepackage{multicol}
\usepackage{algpseudocode,algorithm,algorithmicx}
\usepackage{hyperref}
\usepackage[inline]{enumitem} 
\usepackage{todonotes}
\usepackage{tablefootnote}

\definecolor{purple}{HTML}{7a00a3}
\definecolor{green}{HTML}{00a35f}
\hypersetup{
    colorlinks,
    citecolor=green,
    linkcolor=purple,
}

\pgfplotsset{compat=1.14}

\begin{document}
\title{Relative contributions of Shakespeare and Fletcher in Henry VIII: An Analysis Based on Most Frequent Words and Most Frequent Rhythmic Patterns}
\titlerunning{Relative contributions of Shakespeare and Fletcher in Henry VIII}
\author{Petr Plecháč}
\institute{Institute of Czech Literature, Czech Academy of Sciences, Prague\\
\email{plechac@ucl.cas.cz}}
\maketitle          
\begin{abstract}
The versified play \textit{Henry VIII} is nowadays widely recognized to be a collaborative work not written solely by William Shakespeare. We employ combined analysis of vocabulary and versification together with machine learning techniques to determine which authors also took part in the writing of the play and what were their relative contributions. Unlike most previous studies, we go beyond the attribution of particular scenes and use the rolling attribution approach to determine the probabilities of authorship of pieces of texts, without respecting the scene boundaries. Our results highly support the canonical division of the play between William Shakespeare and John Fletcher proposed by James Spedding, but also bring new evidence supporting the modifications proposed later by Thomas Merriam.
\end{abstract}


\section{Introduction}
In the first collection of William Shakespeare’s works published in 1623 (the so-called \textit{First Folio}) a play appears entitled \textit{The Famous History of the Life of King Henry the Eight} for the very first time. Nowadays it is widely recognized that along with Shakespeare, other authors were involved in the writing of this play, yet there are different opinions as to who these authors were and what the precise shares were of their authorial contributions. This article aims to contribute to the question of the play’s authorship using combined analysis of vocabulary and versification and modern machine learning techniques (as proposed in \cite{plechac2018,plechac2019}).


\section{History and related works}
While the stylistic dissimilarity of \textit{Henry VIII} (henceforth \textit{H8}) to Shakespeare’s other plays had been pointed out before \cite{roderick1758}, it was not until the mid-nineteenth century that Shakespeare’s sole authorship was called into question. In 1850 British scholar James Spedding published an article \cite{spedding1850} attributing several scenes to John Fletcher. Spedding supported this with data from the domain of versification, namely the ratios of iambic lines ending with a stressed syllable (“The view of earthly glory: men might \textit{say}”) to lines ending with an extra unstressed one (“Till this time pomp was single, but now \textit{married}”), pointing out that the distribution of values across scenes is strongly bimodal. 

Since then many scholars have brought new evidence supporting Spedding’s division of the play based both on versification and linguistic features. This includes e.g. frequencies of enjambment \cite{funrivall1874}, frequencies of particular types of unstressed line endings \cite{ingram1874,oras1953}, frequencies of contractions \cite{farnham1916}, vocabulary richness \cite{hart1941}, phrase length measured by the number of words \cite{jackson1997}, or complex versification analysis \cite{tarlinskaja1987,tarlinskaja2014}.\footnote{A detailed history of \textit{H8}’s attributions is given in \cite{vickers2004}.}
From the very beginning, beside advocates of Shakespeare’s sole authorship (e.g. \cite{maxwell1923,alexander1931}), there were also those who supported alternative hypotheses concerning mixed authorship of either Shakespeare, Fletcher, and Philip Massinger \cite{fleay1885,fleay1886,oliphant1891}, Fletcher and Massinger only \cite{boyle1886,sykes1919}, Shakespeare and an unknown author \cite{ege1922}, Shakespeare, Fletcher, Massinger, and an unknown author \cite{merriam1979,merriam1980} or Shakespeare and Fletcher with different shares than those proposed by Spedding \cite{hoy1962}. 

More recent articles usually fall in the last mentioned category and attribute the play to Shakespeare and Fletcher (although the shares proposed by them differ). Thomas Horton \cite{horton1987} employed discriminant analysis of three sets of function words and on this basis attributed most of the scenes to Shakespeare or left them undecided. Thomas Merriam proposed a modification to Spedding’s original attribution concerning re-attribution of several parts of supposedly Fletcher’s scenes back to Shakespeare and vice versa. This was based on measuring the confidence intervals and principal component analysis of frequencies of selected function words in Shakespeare’s and Fletcher’s plays \cite{merriam2003a}, controversial CUSUM technique concerning the occurrences of another set of selected function words and lines ending with an extra unstressed syllable \cite{merriam2003b} or principal component analysis of 64 most frequent words \cite{merriam2018}.\footnote{There were some minor modifications between Merriam’s attribution proposed in \cite{merriam2003a,merriam2003b} and the one in \cite{merriam2018}.} Eisen, Riberio, Segarra, and Egan \cite{eisen2017} used Word adjacency networks \cite{segarra2013} to analyze the frequencies of collocations of selected function words in particular scenes of the play. In contrast to Spedding, they reattribute several scenes back to Shakespeare. Details on Spedding’s attribution as well as the ones mentioned in this paragraph are given in Table \ref{tab:attributions}.

In the present study, with regard to the aforementioned studies, Shakespeare, Fletcher, and Massinger are considered as candidates to the authorship of \textit{H8}.

\begin{table}[ht]
 \center 
 \vspace*{0.3cm}
\begin{tabular}{llccccc}
\hline
  Act &  Scene  &  Spedding\tablefootnote{Originally the prologue and epilogue were left unassigned in \cite{spedding1850}. Their attribution to Fletcher was introduced in the same year in \cite{hickson1850} and was adopted in later works.} & Hoy & Horton & Eisen et al. & Merriam \\
\toprule

\multicolumn{2}{c}{Prologue} & F & N & N & N & F \\      
I & 1 & S & S & S & S & S \\
  & 2 & S & S & S & S & S \\
  & 3 & F & F & S & F & F \\
  & 4 & F & F & F & F & F \\

II & 1 & F & S* & N & S & F \\
 & 2 & F & S* & N & S & $\textrm{S}_{1049} \textrm{F}_{1164} \textrm{S}$ \\
 & 3 & S & S & N & N & $\textrm{S}_{1261} \textrm{F}_{1299} \textrm{S}$ \\
 & 4 & S & S & S & S & S \\
 
III & 1 & F & F & N & F & $\textrm{S}_{1643} \textrm{F}$ \\

 & 2 & $\textrm{S}_{2081} \textrm{F}$ & $\textrm{S}_{2081} \textrm{S*}$ & $\textrm{S}_{2081} \textrm{N}$ & $\textrm{S}$ & $\textrm{S}_{2081} \textrm{F}_{2106} \textrm{S}_{2119} \textrm{F}_{2140} \textrm{S}_{2222} \textrm{F}$ \\

IV & 1 & F & S* & S & S & $\textrm{F}_{2445} \textrm{S}_{2501} \textrm{F}$ \\
 & 2 & F & S* & S & S & $\textrm{F}_{2584} \textrm{S}_{2679} \textrm{F}$ \\

V & 1 & S & S & S & S & S \\
 & 2 & F & F & ? & S & F \\
 & 3 & F & F & ? & S & F \\
 & 4 & F & F & S & N & F \\
 & 5 & F & F & F & F & F \\

\multicolumn{2}{c}{Epilogue} & F & N & N & N & F \\    

    \bottomrule
\end{tabular}
\caption{Selected attributions of \textit{Henry VIII}. \textit{S} denotes attribution of the scene to Shakespeare, \textit{F} denotes Fletcher, \textit{N} denotes unassigned, \textit{S*} denotes Shakespeare with “mere Fletcherian interpolation”. Where the attribution gives precise division of the scene, the subscripted number indicates the last line of a given passage (Through Line Numbering as used in the \textit{Norton Facsimile of the First Folio}).}
    \label{tab:attributions}
\end{table}


\section{Attribution of Particular Scenes}
In the first experiment we perform an attribution of individual scenes of \textit{H8} using the Support Vector Machine\footnote{The implementation of linear SVM in the \textit{scikit-learn} Python library (https://scikit-learn.org/stable/modules/generated/sklearn.svm.SVC.html) have been used.} as a classifier and the frequencies of 500 most frequent rhythmic types\footnote{As a rhythmic type we denote the bit string representing the distribution of stressed and unstressed syllables in a particular line (e.g. “The view of earthly glory: men might say” gives 0101010101, “Till this time pomp was single, but now married” gives 00110100110). Stichomythia are treated as a standalone lines since this approach was found to achieve better accuracy than when half-lines are merged together.} and the frequencies of 500 most frequent words as a feature set. As training samples, individual scenes of plays written by Shakespeare, Fletcher, and Massinger are used\footnote{For both training data and \textit{H8} itself XML versions of the texts provided by \textit{EarlyPrint} project (https://drama.earlyprint.org) have been used. Rhythmic annotation was provided by the \textit{Prosodic} Python library (https://github.com/quadrismegistus/prosodic).} that come roughly from the period when \textit{H8} was supposedly written, namely:

\begin{itemize}
\item Shakespeare: \textit{The Tragedy of Coriolanus} (5 scenes), \textit{The Tragedy of Cymbeline} (27 scenes), \textit{The Winter’s Tale} (12 scenes), \textit{The Tempest} (9 scenes)
\item Fletcher: \textit{Valentinian} (21 scenes), \textit{Monsieur Thomas} (28 scenes), \textit{The Woman’s Prize} (23 scenes), \textit{Bonduca} (18 scenes)
\item Massinger: \textit{The Duke of Milan} (10 scenes), \textit{The Unnatural Combat} (11 scenes), \textit{The Renegado} (25 scenes)
\end{itemize}

Altogether there are thus 53 training samples for Shakespeare, 90 training samples for Fletcher and 46 training samples for Massinger.
In order to estimate the accuracy of the model, cross-validation is performed in the following way:

\begin{itemize}
\item To avoid the risk of overfitting which may be caused by testing the model on the scenes from the same play as it was trained on, we do not perform a standard k-fold cross validation. Instead, we classify scenes of each play by a model trained on the rest, i.e. 5 scenes of Shakespeare’s \textit{Coriolanus} are classified by a model trained on the scenes from the remaining 3 plays by Shakespeare, 4 plays by Fletcher and 5 plays by Massinger, 27 scenes of \textit{Cymbeline} are classified in the same way and so on.
\item Since the training data are imbalanced (which may bias the results), we level the number of training samples per author by random selection.
\item To obtain more representative results, the entire process is repeated 30 times (with a new random selection in each iteration) thus resulting in 30 classifications of each scene.
\item For the sake of comparison of the attribution power of both feature subsets, cross-validations are performed not only of the combined models (500 words $\cup$ 500 rhythmic types), but also of the words-based models (500 words) and versification-based models (500 rhythmic types) alone.
\end{itemize}

As shown in Table \ref{tab:accuracy}, the versification-based models yield a very high accuracy with the recognition of Shakespeare and Fletcher (0.97 to 1 with the exception of Valentinian), yet slightly lower accuracy with the recognition of Massinger (0.81 to 0.88). The accuracy of words-based models remains very high across all three authors (0.95 to 1); in three cases it is nevertheless outperformed by the combined model. We thus may conclude that combined models provide a reliable discriminator between Shakespeare’s, Fletcher’s and Massinger’s styles.

\begin{table}[ht]
 \center 
 \vspace*{0.3cm}
\begin{tabular}{llccc}
\hline
   & &  rhythmic types & words & combination \\
\toprule

Shakespeare & \textit{Coriolanus} & 0.98 & 1 & 1 \\
 & \textit{Cymbeline} & 0.98 & 1 & 1 \\
 & \textit{Winter‘s Tale} & 0.99 & 1 & 1 \\
 & \textit{Tempest}& 0.97 & 1 & 1 \\
Fletcher & \textit{Valentinian} & 0.84 & 0.95 & 0.96 \\
 & \textit{Monsieur Thomas} & 1 & 0.98 & 1 \\
 & \textit{Woman‘s Prize} & 0.98 & 1 & 1 \\
 & \textit{Bonduca} & 0.98 & 0.98 & 1 \\
Massinger & \textit{Duke of Milan} & 0.81 & 0.99 & 0.99 \\
 & \textit{Unnatural Combat} & 0.83 & 1 & 1 \\
 & \textit{Renegado} & 0.88 & 1 & 1 \\

\bottomrule
\end{tabular}
\caption{Accuracy of authorship recognition provided by the models based on (1) 500 most frequent rhythmic types, (2) 500 most frequent words, (3) 1000-dimensional vectors combining features (1) and (2). The number gives the share of correctly classified scenes through all 30 iterations.}
    \label{tab:accuracy}
\end{table}

Table \ref{tab:attrib_h8} gives the results of the classifiers when applied to the individual scenes of \textit{H8}\footnote{The prologue, epilogue and second scene of act 5 (which consists mostly of prose) were not classified due to low number of verse lines.} on the basis of which we may conclude:

\begin{itemize}
\item It is very unlikely that Massinger took part in the text of \textit{H8}. Out of 17 scenes only 2 are attributed to Massinger by any of the models (2.1, 4.2), and in both cases by a mere minority of votes.
\item The probability that the text of \textit{H8} is a result of collaboration between Shakespeare and Fletcher is very high: with 7 scenes all the 30 models agree upon Shakespeare’s authorship, with 5 scenes all the 30 models agree upon Fletcher’s authorship.
\item Our results correspond to the Spedding’s attribution to a high extent. With the exception of two scenes, the majority of models always predict the same author to which it is attributed by Spedding. The two exceptions are the second scene of act 3, where Spedding supposed mixed authorship, and the first scene of act 4, which was originally attributed to Fletcher. 
\end{itemize}

\begin{table}[ht]
 \center 
 \vspace*{0.3cm}
\begin{tabular}{lcccc}
\hline
   & \multicolumn{3}{c}{Classifications results} & \\
\toprule
 & Shakespeare & Fletcher & Massinger & Spedding's attribution \\ 
1.1 & \textbf{30} & 0 & 0 & Shakespeare \\
1.2 & \textbf{30} & 0  & 0 & Shakespeare \\
1.3 & 0 & \textbf{30} & 0 & Fletcher \\
1.4 & 0 & \textbf{30} & 0 & Fletcher \\
2.1 & 0 & \textbf{20} & 10 & Fletcher \\
2.2 & 0 & \textbf{30} & 0 & Fletcher \\
2.3 & \textbf{30} & 0 & 0 & Shakespeare \\
2.4 & \textbf{30} & 0 & 0 & Shakespeare \\
3.1 & 0 & \textbf{30} & 0 & Fletcher \\
3.2 & \textbf{30} & 0 & 0 & \textbf{Shakespeare/Fletcher} \\
4.1 & \textbf{30} & 0 & 0 & \textbf{Fletcher} \\
4.2 & 0 & \textbf{23} & 7 & Fletcher \\
5.1 & \textbf{30} & 0 & 0 & Shakespeare \\
5.3 & 9 & \textbf{21} & 0 & Fletcher \\
5.4 & 7 & \textbf{23} & 0 & Fletcher \\
5.5 & 0 & \textbf{30} & 0 & Fletcher \\

\bottomrule
\end{tabular}
\caption{Classification of individual scenes of \textit{H8}. The number indicates how many times out of 30 iterations the author has been predicted to a given scene. The highest value in each row is printed in bold. The rightmost column indicates to which author the scene is attributed by Spedding. Where Spedding differs from our results, we use a bold face.}
    \label{tab:attrib_h8}
\end{table}


\section{Rolling attribution of the play}
Even though the classification of individual scenes clearly indicates that \textit{H8} is a result of collaboration between Shakespeare and Fletcher, we should not accept it as the final result since most studies suggest that—at least in the case of the second scene of act 3—the shift of authorship did not happen on the scenes’ boundaries (as shown in Table  \ref{tab:attributions}). To get a more reliable picture of the authors’ shares, we’ve employed so called \textit{rolling attribution}.

Rolling attribution was originally introduced by Maciej Eder \cite{eder2016} as a technique designed for cases involving mixed authorship. Unlike common tasks, in rolling attribution neither the entire text nor its logical parts (chapters, scenes etc.) are being classified but instead its overlapping parts of fixed length. Assume a text which one supposes to be a result of a collaboration between two (or more) authors consisting of $n$ lines $l_1, l_2, l_3, \ldots, l_{n}$. Let $k$ and $d$ be arbitrarily chosen values so that $k \in \mathbb{N}$, $k < n$ and $d \in \mathbb{N}$, $d < n - k$, $d \leq k$. For each $i; i \in \{0, d, 2d, 3d, \ldots\}, i < n - k$ a battery of attributions is performed of all the sections s consisting of lines $l_{i+1}, l_{i+2}, l_{i+3}, \ldots, l_{i+k}$.\footnote{In Eder’s original article the length of sections is measured by a number of words not lines.} To achieve a better sensitivity to authorship transitions Eder suggests not to work with simple predictions (labeling the section as being written by a single author) but—if it’s possible with a given classifier—rather a probability distribution over candidate authors.

We first test the performance of rolling attribution on 4 plays by Shakespeare and 4 plays by Fletcher contained in the training set. For each play we train 30 models on the remaining data with number of training samples randomly leveled in each iteration. Each target play is segmented into overlapping parts with $k = 100$ and $d = 5$ (each successive series of five lines except for the initial 19 and final 19 ones are thus classified 600 times—30 times within 20 different parts). The output of classification of each part is transformed to probability distribution using Platt’s scaling \cite{platt1999}.

\begin{figure}
\begin{subfigure}{.5\textwidth}
  \centering
  \includegraphics[width=1\linewidth]{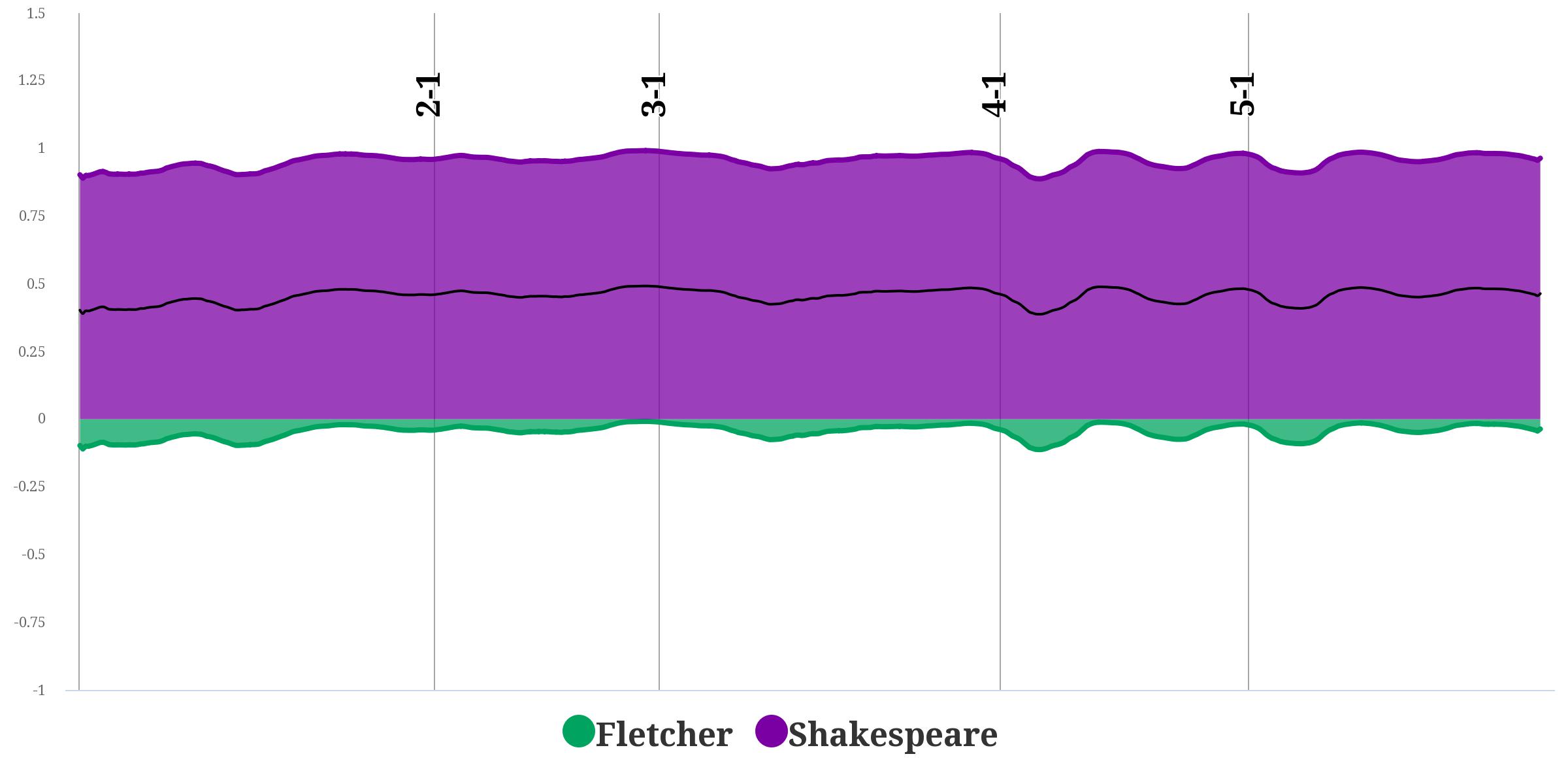}
  \caption{Shakespeare: \textit{Coriolanus}}
  \label{fig:sfig1}
\end{subfigure}%
\begin{subfigure}{.5\textwidth}
  \centering
  \includegraphics[width=1\linewidth]{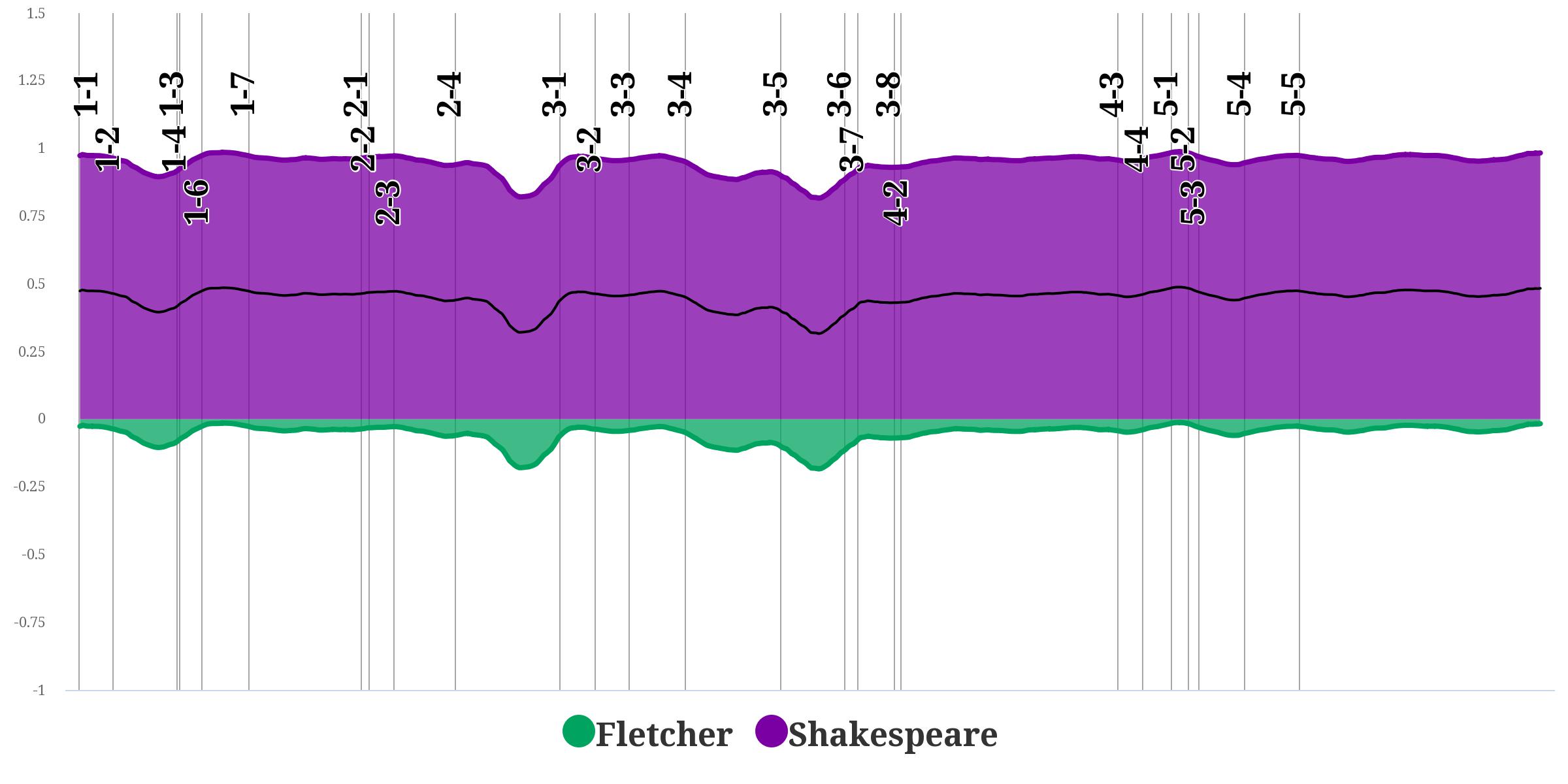}
  \caption{Shakespeare: \textit{Cymbeline}}
  \label{fig:sfig2}
\end{subfigure}

\begin{subfigure}{.5\textwidth}
  \centering
  \includegraphics[width=1\linewidth]{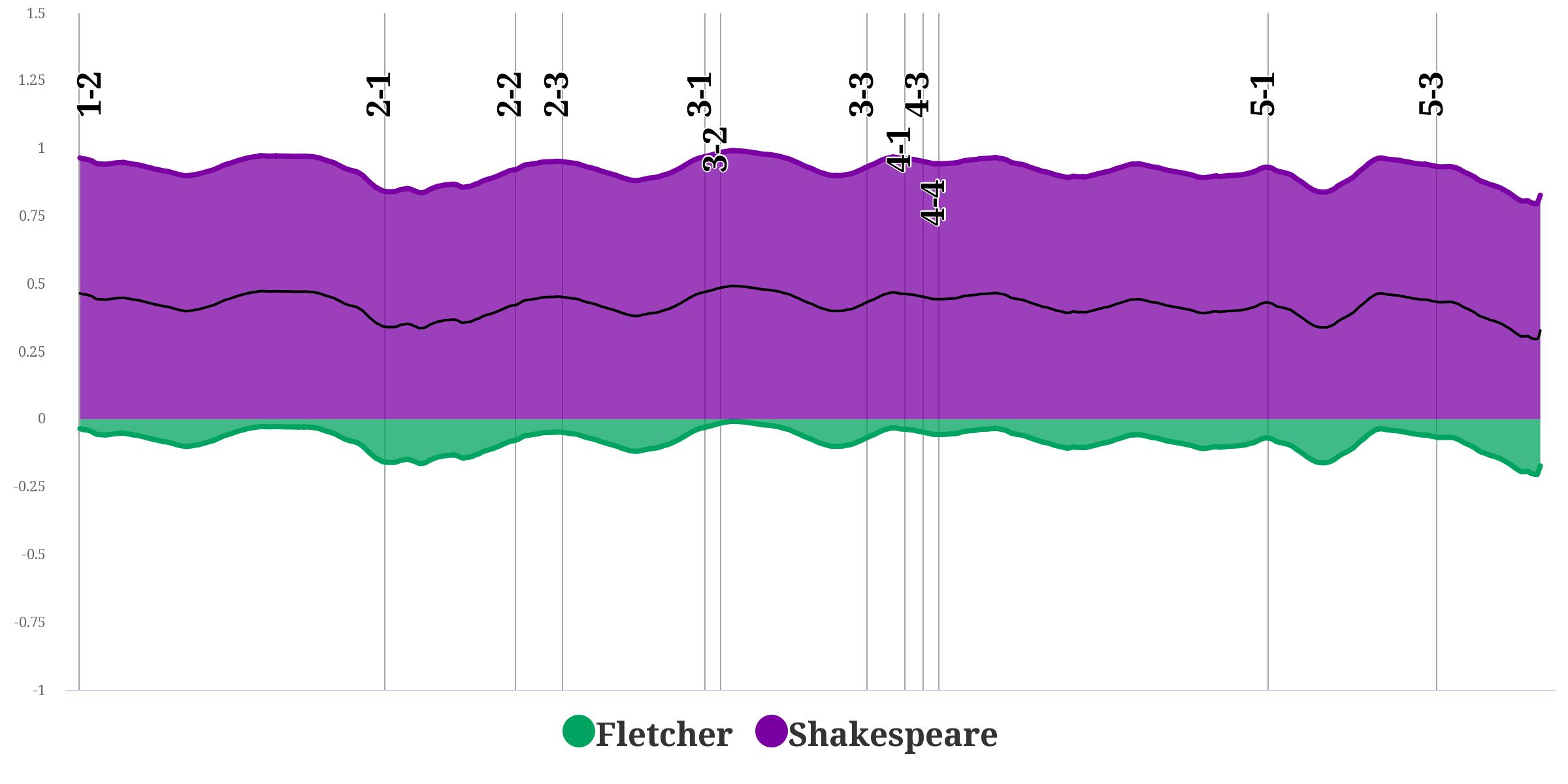}
  \caption{Shakespeare: \textit{Winter's Tale}}
  \label{fig:sfig3}
\end{subfigure}%
\begin{subfigure}{.5\textwidth}
  \centering
  \includegraphics[width=1\linewidth]{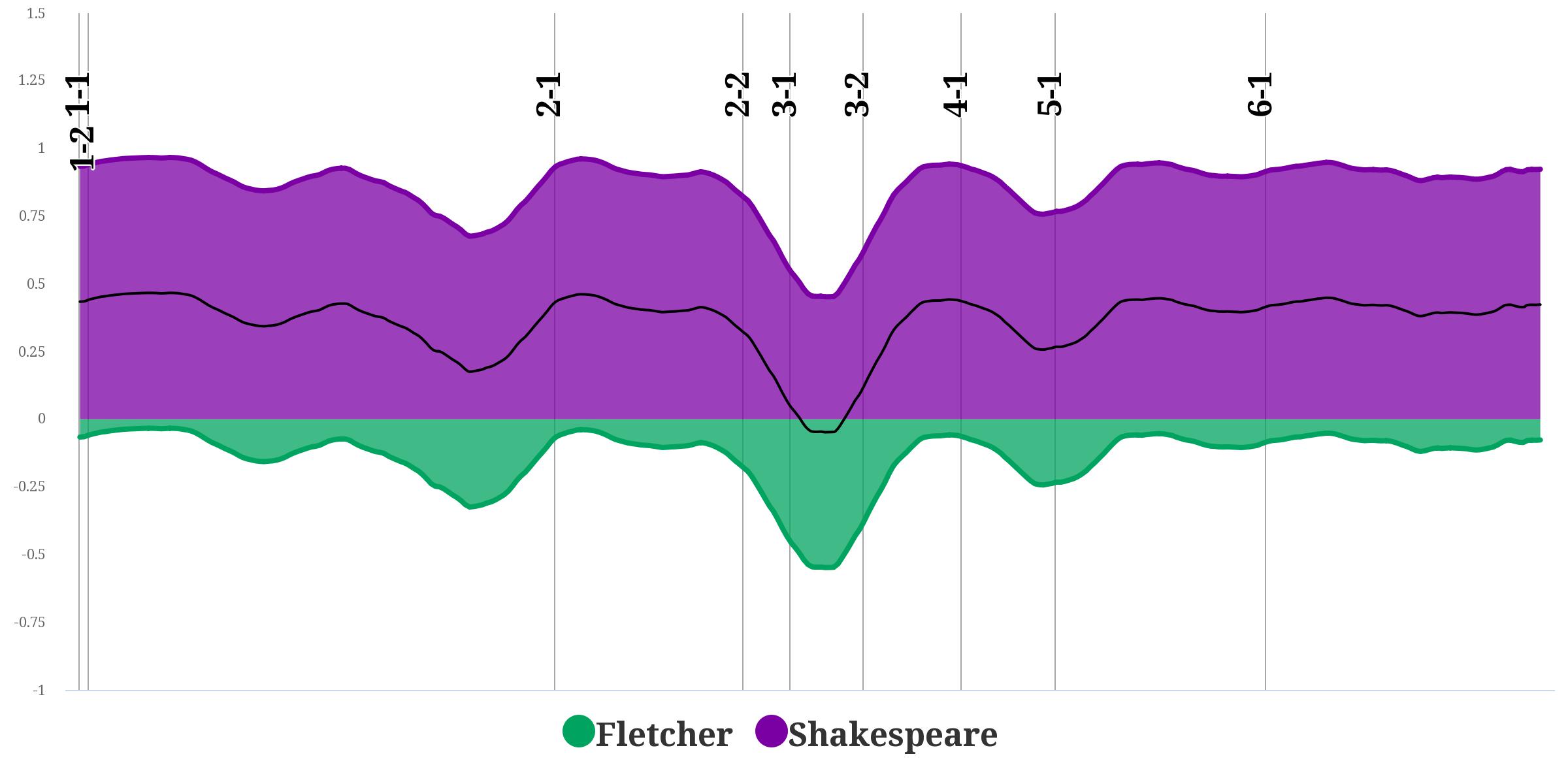}
  \caption{Shakespeare: \textit{Tempest}}
  \label{fig:sfig4}
\end{subfigure}

\begin{subfigure}{.5\textwidth}
  \centering
  \includegraphics[width=1\linewidth]{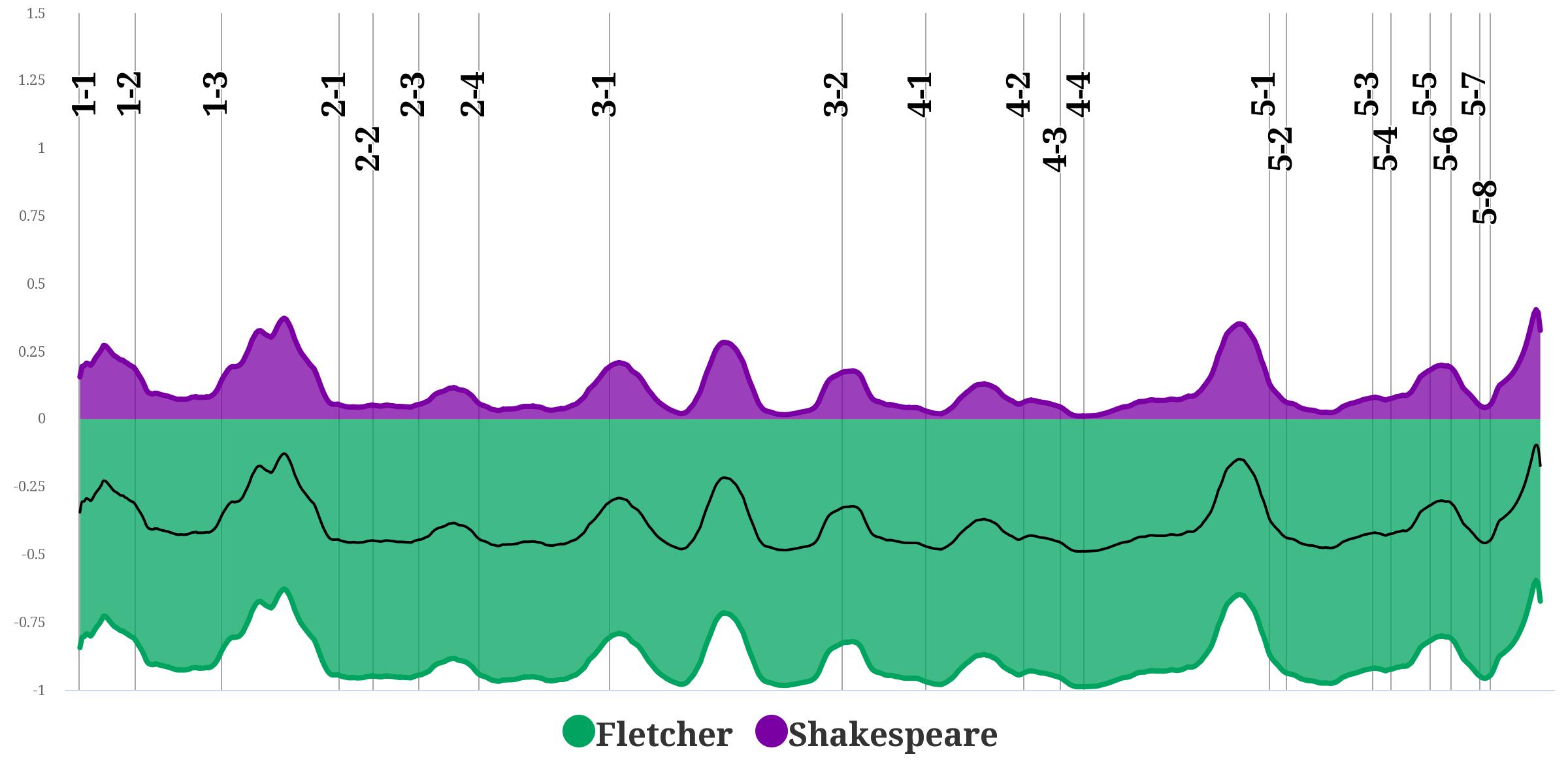}
  \caption{Fletcher: \textit{Valentinian}}
  \label{fig:sfig5}
\end{subfigure}%
\begin{subfigure}{.5\textwidth}
  \centering
  \includegraphics[width=1\linewidth]{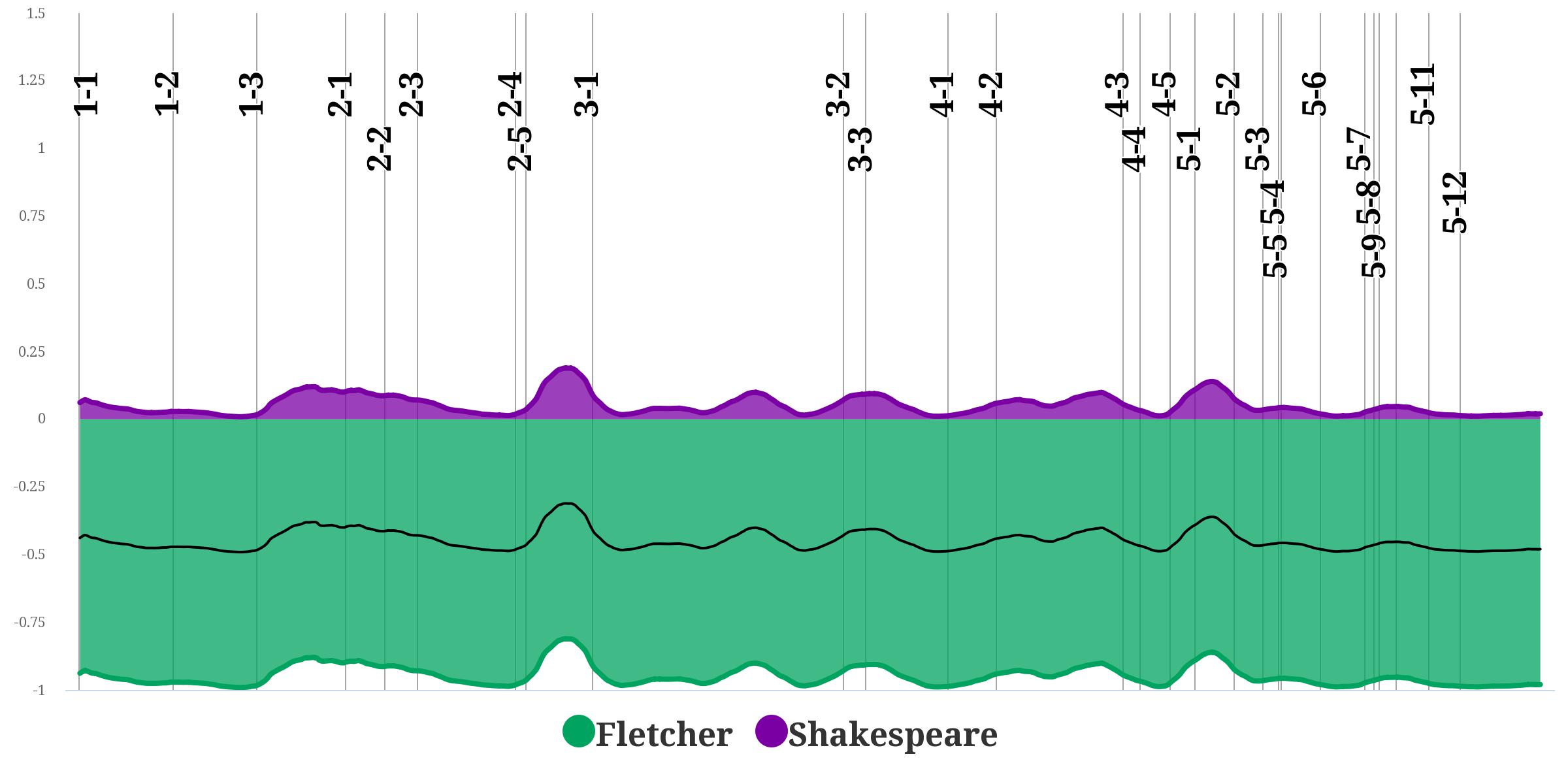}
  \caption{Fletcher: \textit{Monsieur Thomas}}
  \label{fig:sfig6}
\end{subfigure}

\begin{subfigure}{.5\textwidth}
  \centering
  \includegraphics[width=1\linewidth]{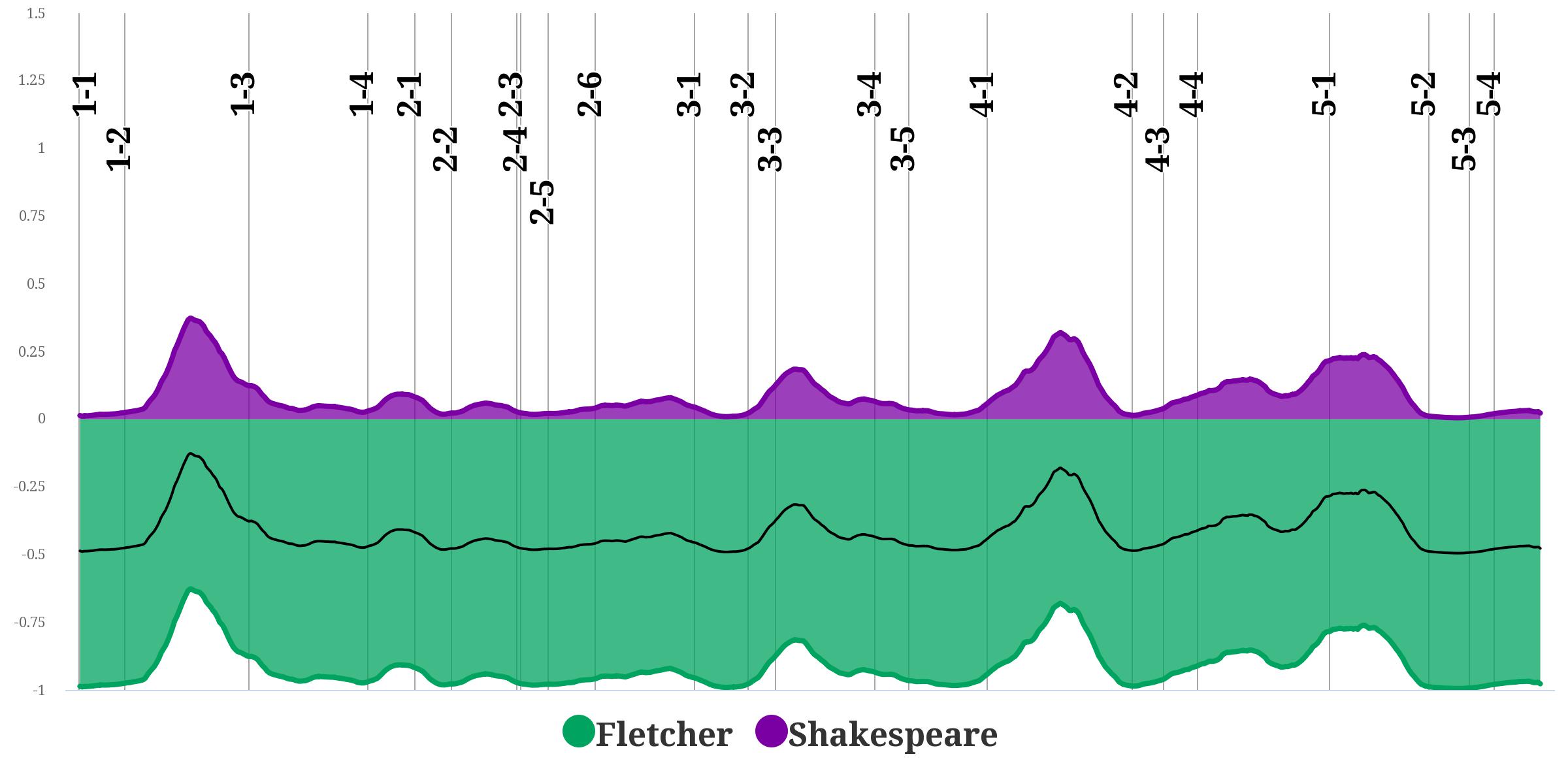}
  \caption{Fletcher: \textit{Woman's Prize}}
  \label{fig:sfig7}
\end{subfigure}%
\begin{subfigure}{.5\textwidth}
  \centering
  \includegraphics[width=1\linewidth]{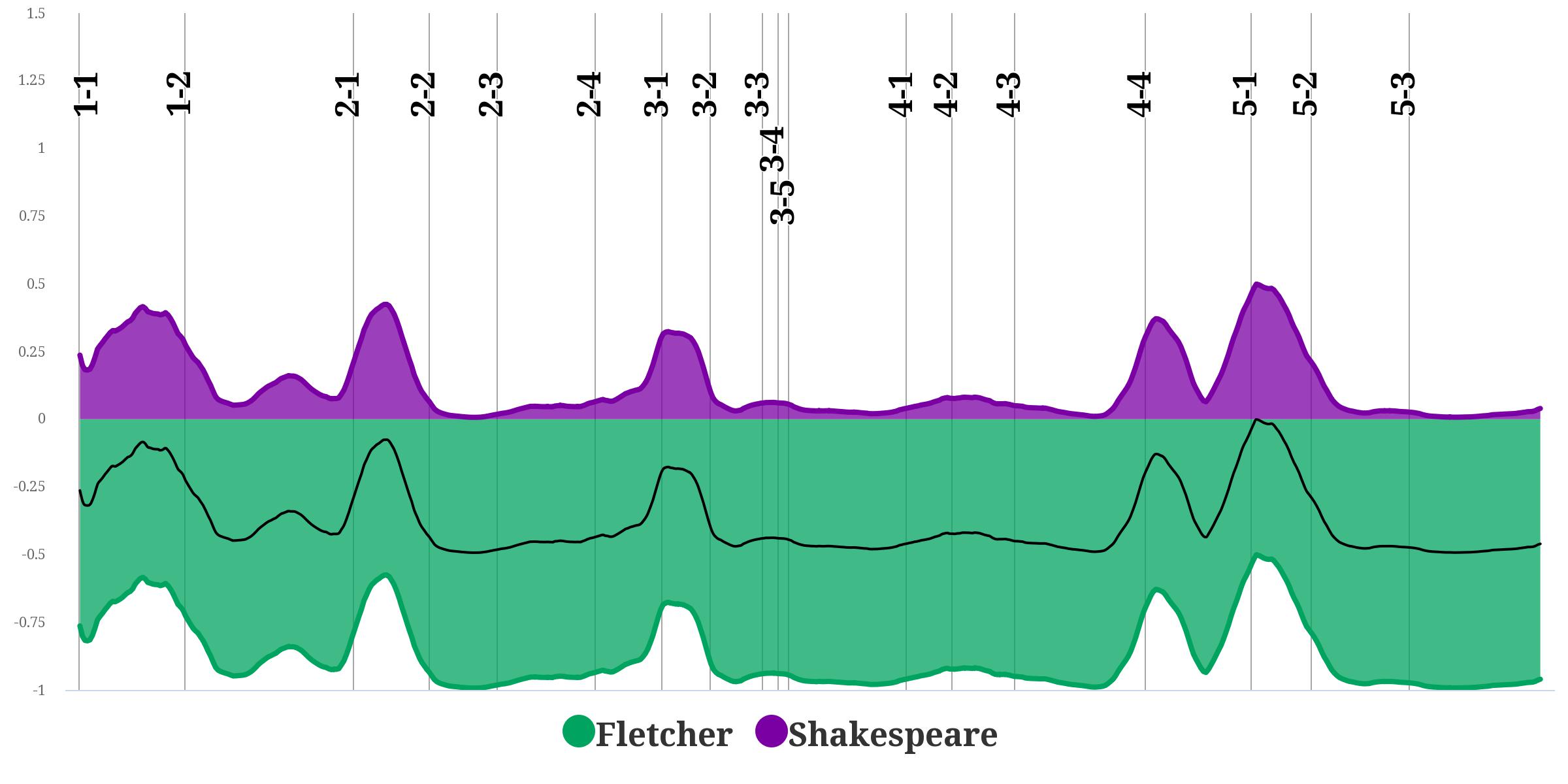}
  \caption{Fletcher: \textit{Bonduca}}
  \label{fig:sfig8}
\end{subfigure}

\caption{Rolling attribution of 4 plays by Shakespeare and 4 plays by Fletcher based on 500 most frequent rhythmic types and 500 most frequent words. Vertical lines indicate scene boundaries.}
\label{fig:rolling_train}
\end{figure}

Fig. \ref{fig:rolling_train} gives the results for each of the eight plays. Each data point corresponds to a group of five lines and gives the mean probability of Shakespeare’s and Fletcher’s authorship. For the sake of clarity, the values for Fletcher are displayed as negative. The distance between Shakespeare’s data point and Fletcher’s data point thus always equals 1. The black curve gives the average of both values. The results suggest the rolling attribution method with combined versification and lexical features to be very reliable: (1) Probability of Fletcher’s authorship is very low for vast majority of Shakespeare’s work. The only place where Fletcher is assigned higher probability than Shakespeare is the sequence of 10 five-line groups in the second act of scene 2 of the \textit{Tempest}. (2) Probability of Shakespeare’s authorship is very low for vast majority of Fletcher’s work. The only place where Shakespeare comes closer to Fletcher’s values is the first scene of act 5 of \textit{Bonduca}. Having only 10 groups misattributed out of 4412 we may estimate the accuracy of rolling attribution to be as high as 0.9977 when distinguishing between Shakespeare and Fletcher.

After validation of the method we proceed to \textit{H8}. Fig. \ref{fig:rolling_henry} gives the results of rolling attribution based on a combined vector of most frequent types and most frequent words, and additionally for each of these feature subsets alone. Models were trained on all 8 plays in the training set with the same setting as above ($k = 100; d = 5$). It once again supports Spedding’s attribution to a high extent:

\begin{figure}[!htb]
  \centering
      \includegraphics[width=1\textwidth]{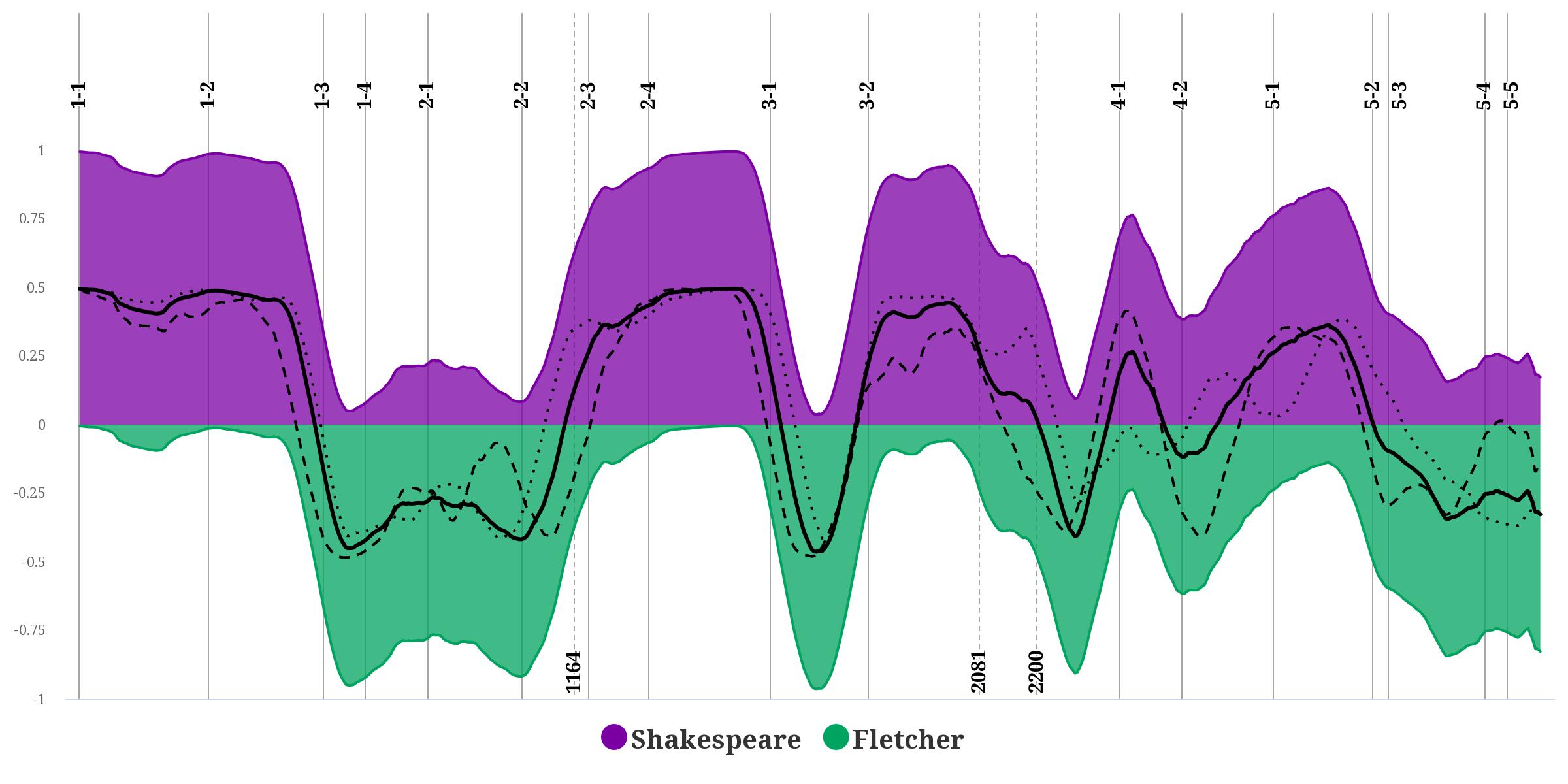}
  \caption{Rolling attribution of \textit{H8} based on 500 most frequent rhythmic types and 500 most frequent words. Vertical lines indicate scene boundaries (label on top) or other landmark indicated in other articles (label on bottom giving the line number according to TLN as used in the \textit{Norton Facsimile of the First Folio}). Dashed line indicates results of rolling attribution based solely on 500 most frequent rhythmic types, dotted line indicates results of rolling attribution based solely on 500 most frequent words.}
  \label{fig:rolling_henry}
\end{figure}

\begin{itemize}
\item For scenes \textbf{1.1} and \textbf{1.2} rhythmic types, words as well as the combined model indicate Shakespeare to be the author. All three sets of models indicate that the shift of authorship happened at the end of scene 1.2.
\item For scenes \textbf{1.3, 1.4, 2.1} and \textbf{2.2} all three sets of models indicate Fletcher to be the author. Rhythmic types indicate that the shift of authorship happened at the end of 2.2, while word-based models indicate that the shift happened before the end of the scene. (Recall that the shift of authorship within 2.2 is proposed also by Thomas Merriam (cf. Table \ref{tab:attributions}) even though a little bit further at line 1164.)
\item Scenes \textbf{2.3} and \textbf{2.4} are according to all sets of models authored by Shakespeare. All three sets of models indicate that the shift happened at the end of scene 2.4.
\item According to all sets of models, scene \textbf{3.1} was written by Fletcher. All three sets of models indicate that the shift happened at the scene’s end.
\item Scene \textbf{3.2} is usually attributed to both Shakespeare and Fletcher. All three sets of models support this. While Spedding and other authors locate the shift to line 2081, all our sets of models indicate that it occurred later. Combined models locate it precisely at line 2200 (in agreement with earlier studies by Merriam \cite{merriam2003a,merriam2003b}. A certain decrease in the probability of Shakespeare’s authorship found in the neighborhood of line 2081 in word-based models and combined models may support Merriam’s later attributions \cite{merriam2018}, i.e. mixed authorship even after the line 2081.
\item For scenes \textbf{4.1} and \textbf{4.2} the rhythmic types indicate Shakespeare’s authorship of the first (contrary to Spedding) and Fletcher’s authorship of the latter. Location of the shift does not however fully correspond to the scene boundaries. Probabilities extracted from word-based models and combined models are close to 0.5 for both authors which may support Merriam’s attribution (mixed authorship).
\item Scene \textbf{5.1} is according to all sets of models authored by Shakespeare. Rhythmic types and combined models locate the shift at its end; word-based models locate it a little later on.
\item Scenes \textbf{5.2, 5.3, 5.4} and \textbf{5.5} are Fletcher’s according to word-based models and combined models. Rhythmic types indicate the possibility of Shakespeare’s share in 5.4.
\end{itemize}


\section{Conclusions}
Combined versification-based and word-based models trained on 17th century English drama yield a high accuracy of authorship recognition. We can thus state with high reliability that \textit{H8} is a result of collaboration between William Shakespeare and John Fletcher, while the participation of Philip Massinger is rather unlikely.

The rolling attribution method suggests that particular scenes are indeed mostly a work of a single author and that their contributions roughly correspond to what has been proposed by James Spedding \cite{spedding1850}. The main differences between our results and Spedding’s attribution are the ambivalent outputs of models for both scenes of act 4. However, it is worth noting that Spedding himself expressed some doubts about the authorship of these scenes.\footnote{“Of the 4th Act I did not so well know what to think. For the most part it seemed to bear evidence of a more vigorous hand than Fletcher‘s, with less mannerism, especially in the description of the coronation, and the character of Wolsey; and yet it had not to my mind the freshness and originality of Shakspere” \cite{spedding1850}.} Other differences are rather marginal and usually support the modifications of Spedding’s original attribution, as proposed by Thomas Merriam \cite{merriam2003a,merriam2003b,merriam2018}.

\bibliographystyle{ieeetr}
\bibliography{biblio}
\end{document}